\documentclass[letterpaper]{article} 
\usepackage{aaai25}  
\usepackage{times}  
\usepackage{helvet}  
\usepackage{courier}  
\usepackage[hyphens]{url}  
\usepackage{graphicx} 
\usepackage{natbib}  
\usepackage{caption} 
\usepackage{amsmath}
\usepackage{amssymb}
\usepackage{verbatim}
\usepackage{booktabs}
\usepackage{multirow}
\usepackage{makecell}

\usepackage{tcolorbox}

\DeclareUnicodeCharacter{0301}{\'{e}}

\usepackage{colortbl}
\definecolor{yan}{rgb}{0.5,0.3,0.5}

\definecolor{ben}{rgb}{0.9,0.,0.5}

\definecolor{navyblue}{RGB}{191, 209, 229} 
\definecolor{light_yellow}{RGB}{255,243,194}
\definecolor{orange}{RGB}{255,200,100}
\definecolor{red}{RGB}{255, 0, 0}
\definecolor{green}{RGB}{0, 176, 80}
\newcommand{\best}{\cellcolor{orange}}

\usepackage{algorithm}
\usepackage{algorithmic}

\usepackage{newfloat}
\usepackage{listings}
\DeclareCaptionStyle{ruled}{labelfont=normalfont,labelsep=colon,strut=off} 
\floatstyle{ruled}
\newfloat{listing}{tb}{lst}{}
\floatname{listing}{Listing}
\title{MMGDreamer: Mixed-Modality Graph for Geometry-Controllable \\ 3D Indoor Scene Generation}
\author{
    Zhifei Yang\textsuperscript{\rm 1}\thanks{Work done at Beijing Digital Native Digital City Research Center.}, 
    Keyang Lu\textsuperscript{\rm 2},
    Chao Zhang\textsuperscript{\rm 3}\footnote{Corresponding author.},
    Jiaxing Qi\textsuperscript{\rm 2},
    Hanqi Jiang\textsuperscript{\rm 3},
    Ruifei Ma\textsuperscript{\rm 3},
    Shenglin Yin\textsuperscript{\rm 1},
    Yifan Xu\textsuperscript{\rm 2},
    Mingzhe Xing\textsuperscript{\rm 1}, 
    Zhen Xiao\textsuperscript{\rm 1}\footnotemark[2],
    Jieyi Long\textsuperscript{\rm 4},
    Guangyao Zhai\textsuperscript{\rm 5}
}
\affiliations{
    \textsuperscript{\rm 1} School of Computer Science, Peking University \\
    \textsuperscript{\rm 2} Beihang University\\
    \textsuperscript{\rm 3} Beijing Digital Native Digital City Research Center \\
    \textsuperscript{\rm 4} Theta Labs, Inc. \\
    \textsuperscript{\rm 5} Technical University of Munich\\[0.5em]
    yangzhifei@stu.pku.edu.cn, lukeyang@buaa.edu.cn, ariczhang2009@gmail.com, \\ 
    xiaozhen@pku.edu.cn, jieyi@thetalabs.org, guangyao.zhai@tum.de

}

\usepackage{bibentry}

\begin{document}

\maketitle

\begin{abstract}
Controllable 3D scene generation has extensive applications in virtual reality and interior design, where the generated scenes should exhibit high levels of realism and controllability in terms of geometry. Scene graphs provide a suitable data representation that facilitates these applications.
However, current graph-based methods for scene generation are constrained to text-based inputs and exhibit insufficient adaptability to flexible user inputs, hindering the ability to precisely control object geometry.
To address this issue, we propose \textbf{MMGDreamer}, a dual-branch diffusion model for scene generation that incorporates a novel \textbf{Mixed-Modality Graph}, visual enhancement module, and relation predictor. 
The mixed-modality graph allows object nodes to integrate textual and visual modalities, with optional relationships between nodes. It enhances adaptability to flexible user inputs and enables meticulous control over the geometry of objects in the generated scenes.
The visual enhancement module enriches the visual fidelity of text-only nodes by constructing visual representations using text embeddings.
Furthermore, our relation predictor leverages node representations to infer absent relationships between nodes, resulting in more coherent scene layouts.
Extensive experimental results demonstrate that MMGDreamer exhibits superior control of object geometry, achieving state-of-the-art scene generation performance.

\end{abstract}

\begin{links}
    \link{Project page}{https://yangzhifeio.github.io/project/MMGDreamer}
\end{links}
%

\begin{figure*}[t]
\centering
\includegraphics[width=1\textwidth]{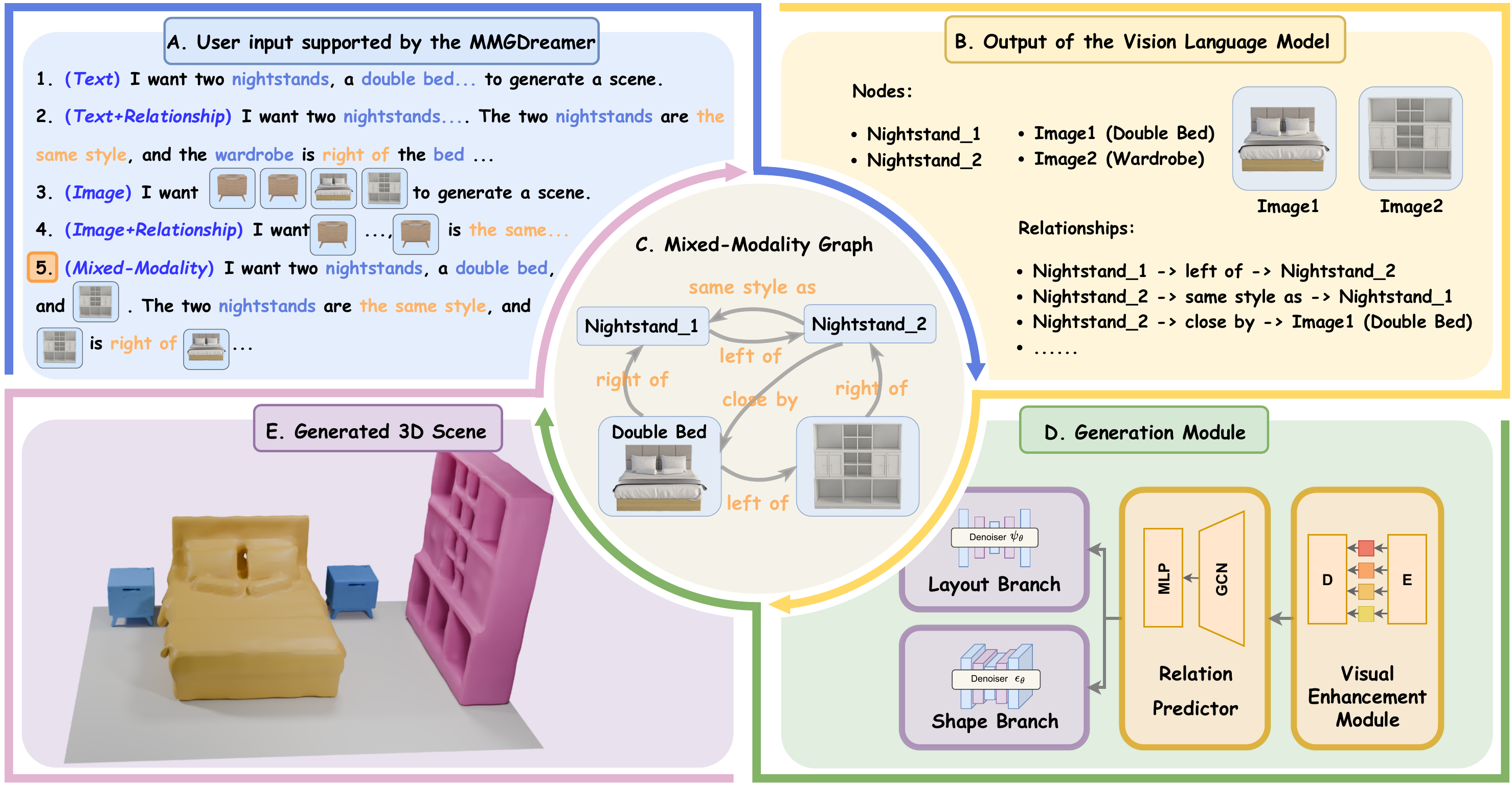} 
\caption{\textbf{MMGDreamer} processes a Mixed-Modality Graph to generate a 3D indoor scene, where object geometry can be precisely controlled. Starting from the fifth type of input (Mixed-Modality) shown in module A as an example, the framework utilizes a vision-language model (B) to produce a Mixed-Modality Graph (C). This graph is further refined by the Generation Module (D) to create a coherent and precise 3D scene (E).}
\label{introduction}
\end{figure*}

\section{Introduction}
Deep generative models have initiated a new era of artificial intelligence-generated content, driving developments in natural language generation \cite{zheng2024judging}, video synthesis \cite{liao2024videoinsta}, and 3D generation \cite{poole2022dreamfusion}. 
Controllable Scene Generation refers to generating realistic 3D scenes based on input prompts, allowing for precise control and adjustment of specific objects within those scenes. 
It is widely applied in Virtual Reality \cite{bautista2022gaudi}, Interior Design \cite{ ccelen2024design}, and Embodied Intelligence\cite{yang2024physcene, zhai2024sg}, providing immersive experiences and enhancing decision-making processes.
Within these applications, scene graphs serve as a powerful tool by succinctly abstracting the scene context and interrelations between objects, enabling intuitive scene manipulation and generation \cite{dhamo2021graph}.

Despite the significant progress made by retrieval-based \cite{lin2024instructscene}, semi-generative \cite{ren2024xcube}, and fully-generative \cite{zhai2024commonscenes} methods in graph-based controllable scene generation, these approaches predominantly rely on textual descriptions to construct input scene graphs. While text serves as a high-level representation encapsulating rich semantic information, it falls short in accurately describing the geometry of objects in the generated scenes, resulting in inadequate geometric control over the generated objects \cite{rombach2022high}. 
Moreover, each node in the scene graph contains only textual information about object category, which limits its adaptability to flexible user input.
To address these limitations, we introduce MMGDreamer, a dual-branch diffusion model designed for processing multimodal information, incorporating a novel Mixed-Modality Graph (MMG) as a key component. 
As depicted in Fig.~\ref{introduction}, the node of MMG can be represented in three ways: text, image, or a combination of both. 
Additionally, edges between nodes can be selectively provided or omitted based on user input.
This flexible graph structure supports five types of user input, as illustrated in Fig.~\ref{introduction}.A, significantly enhancing adaptability to diverse user demands and enabling precise control over object geometry in generated scenes.

To fully leverage the capabilities of MMG, MMGDreamer features two pivotal modules: the visual enhancement module and the relation predictor. 
When nodes of the input scene graph contain solely textual information, the visual enhancement module employs text embeddings to construct visual representations of these nodes. 
By incorporating visual priors associated with the text, this approach enriches the visual content of nodes, enhancing geometric control over the generated objects. 
The relation predictor, a relationship classifier based on the GCN, leverages prior knowledge and node representations within the scene to infer relationships between nodes in the absence of explicit relational information. By capturing global and local scene-object relationships, this module ensures the generation of more coherent and contextually appropriate scene layouts. 
 We briefly summarize our primary contributions as follows:
\begin{itemize}
    \item We introduce a novel \textbf{Mixed-Modality Graph}, where nodes can selectively incorporate textual and visual modalities, allowing for precise control over the object geometry of the generated scenes and more effectively accommodating flexible user inputs. 
    \item We present \textbf{MMGDreamer}, a dual-branch diffusion model for scene generation based on Mixed-Modality Graph, which incorporates two key modules: a visual enhancement module and a relation predictor, dedicated to construct node visual features and predict relations between nodes, respectively.
    \item Extensive experiments on the SG-FRONT dataset demonstrate that MMGDreamer attains higher fidelity and geometric controllability, and achieves state-of-the-art performance in scene synthesis, outperforming existing methods by a large margin.
\end{itemize}

\section{Related Work}
\subsubsection{Scene Graph.}
Scene graph provides a structured and hierarchical representation of complex scenes by using nodes (objects) and edges (relationships) \cite{zhou2019scenegraphnet}. 
Following their introduction, subsequent works have refined hierarchical scene graph \cite{rosinol20203d} and focused on predicting local inter-object relationships \cite{koch2024lang3dsg,liao2024gentkg}.
Such advancements have driven the widespread application of scene graphs across both 2D and 3D domains, enabling sophisticated tasks such as image synthesis \cite{johnson2018image, wu2023scene} and caption generation \cite{basioti2024cic} in 2D, as well as video synthesis \cite{cong2023ssgvs}, 3D scene understanding \cite{wald2020learning} and scene synthesis \cite{para2021generative} in 3D. 
However, in the current 3D indoor scene synthesis tasks, scene graphs are predominantly derived from text input by users \cite{strader2024indoor}. We propose the MMG, which more effectively accommodates flexible user inputs.

\subsubsection{3D Scene Generation.}
3D scene generation is an area of ongoing research that focuses on developing plausible layouts \cite{engelmann2021points} and generating accurate object shapes \cite{xu2023discoscene}.
A substantial body of contemporary research synthesizes scenes from text \cite{fang2023ctrl}, panorama \cite{wang2023perf, hara2024magritte}, or spatial layout \cite{yan2024frankenstein, jyothi2019layoutvae} using autoregressive paradigms \cite{wang2021sceneformer}, object decoupling techniques \cite{zhang2024style, epstein2024disentangled}, or prior learning \cite{hollein2023text2room}.
In particular, CommonScenes \cite{zhai2024commonscenes} utilizes scene graphs as conditions and adopts an end-to-end framework to generate both object shapes and scene layouts simultaneously. 
EchoScene \cite{zhai2024echoscene} advances the CommonScenes by incorporating an information echo scheme.
Nevertheless, these approaches are inadequate for effectively controlling object geometry. To overcome this limitation, we introduce MMGDreamer, which fully leverages the MMG to achieve precise control over object geometry.


\section{Preliminary}
 
\subsection{Scene Graph Representation}
Scene graph can be formally defined as a directed graph $ \mathcal{G} = \{\mathcal{V}, \mathcal{R}\} $, where $ \mathcal{V} = \{v_i \ | \  i \in \{1, \ldots, N \} \} $ represents the set of nodes (objects in the scene) and $ \mathcal{R} = \{r_{i \to j} \ | \ i, j \in \{1, \ldots, N\}, i \neq j\} $ represents the set of edges (relationships between objects).
The node $v_i$ represents an object in the scene and contains information about the object's category. The edges $ r_{i \to j} $ define the connections between objects, which can include spatial relationships (e.g., front/behind) or attribute relationships (e.g., same style).

\subsection{Scene Graph Encoder}
Graph Convolutional Network (GCN) \cite{johnson2018image} facilitates the processing of graph-structured data by learning node representations through the layer-by-layer aggregation of features from neighboring nodes.
In our work, we utilize the Triplet Graph Convolutional Network (Triplet-GCN) as scene graph encoder to process the scene graph, assuming that the initial node and edge attribute features are given by $(\delta^{(0)}_{v_i}, \delta^{(0)}_{r_{i \to j}}, \delta^{(0)}_{v_j})$.
Specifically, each layer of GCN updates the node and edge representations according to the following formula:
\begin{equation}
(\gamma^{l}_{v_i}, \delta^{l+1}_{r_{i \to j}}, \gamma^{l}_{v_j}) = \text{MLP}_1(\delta^{l}_{v_i}, \delta^{l}_{r_{i \to j}}, \delta^{l}_{v_j}), 
\end{equation}
\begin{equation}
\delta^{l+1}_{v_i} = \gamma^{l}_{v_i} + \text{MLP}_2\left(\text{Avg}(\gamma^{l}_{v_j} \mid v_j \in \mathcal{N}_G(v_i))\right),
\end{equation}
where $l \in \{1, \ldots, L-1\}$ denotes an independent layer within the Triplet-GCN, $\mathcal{N}_G(v_i)$ represents the set of all neighboring nodes of $v_i$, Avg indicates the use of average pooling operation, and $\text{MLP}_1$ and $\text{MLP}_2$ refer to the Multi-Layer Perceptron (MLP) layers. 
\begin{figure*}[t]
\centering
\includegraphics[width=1\textwidth]{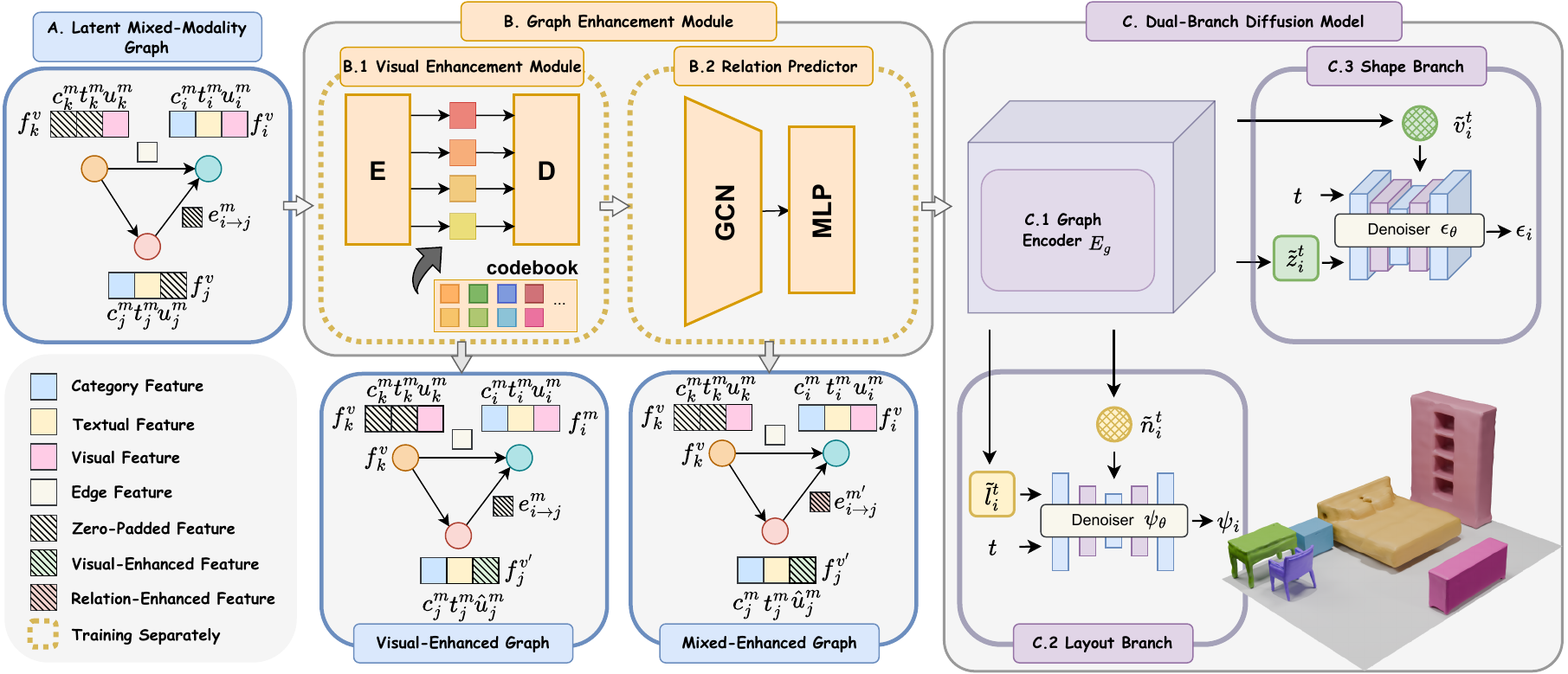} 
\caption{\textbf{Overview of MMGDreamer.} Our pipeline consists of the Latent Mixed-Modality Graph, the Graph Enhancement Module, and the Dual-Branch Diffusion Model. 
During inference, MMGDreamer initiates with the Latent Mixed-Modality Graph, which undergoes enhancement via the Visual Enhancement Module and the Relation Predictor, resulting in the formation of a Visual-Enhanced Graph and a Mixed-Enhanced Graph. 
The Mixed-Enhanced Graph is then input into the Graph Encoder $E_g$ within the Dual-Branch Diffusion Model for relationship modeling, using a triplet-GCN structured module integrated with an echo mechanism.
Subsequently, the Layout Branch (C.2) and the Shape Branch (C.3) use denoisers conditioned on the nodes' latent representations to generate layouts and shapes, respectively.
The final output is a synthesized 3D indoor scene where the generated shapes are seamlessly integrated into the generated layouts.}
\label{pipeline}
\end{figure*}

\subsection{Latent Diffusion Model}
Latent Diffusion Model (LDM) \cite{rombach2022high} generally involves two Markov processes: a forward process that incrementally corrupts the data and a reverse process that progressively denoises it. 
Given a sample $x_0$, the LDM first employs a pre-trained VQ-VAE \cite{van2017neural} to encode $ x_0 $ into a reduced-dimensional latent representation $ z_0$.
The forward process is defined by:
\begin{equation}
q(z_t \mid z_{t-1}) = \mathcal{N}(z_t; \sqrt{1 - \beta_t} z_{t-1}, \beta_t \mathbf{I}),
\end{equation}
where $ t $ ranges from $ 1 $ to $ T $, with $ T $ denoting the total number of timesteps. The parameter $ \beta_t $ controls the noise level introduced at each timestep $ t $. 
In the reverse process, $ z_t $ is processed through a denoiser $ \epsilon_{\theta} $, such as UNet \cite{ronneberger2015u}, to estimate the noise, enabling a progressive denoising process to recover a clean latent representation. The objective function can be written as:
\begin{equation}
    \mathcal{L}_{LDM} = \mathbb{E}_{z_t, t, \epsilon \sim \mathcal{N}(0, 1)} \left[ \| \epsilon - \epsilon_{\theta}(z_t, t, c) \|^2_2 \right],
\end{equation}
where $c$ is a condition to guide the reverse process.

\section{Method}
We propose MMGDreamer, a framework adept at handling MMG as input for indoor scene synthesis tasks, as illustrated in Fig.~\ref{pipeline}. 
The MMG is a novel graph structure where nodes can optionally carry textual or visual information, thereby more effectively accommodating flexible user inputs.
MMGDreamer first utilizes CLIP and an embedding layer to encode the MMG, producing the Latent Mixed-Modality Graph (LMMG). We then apply the visual enhancement module to construct visual information in the nodes of the LMMG, yielding a Visual-Enhanced Graph. Subsequently, a Relation Predictor is utilized to predict the missing edges between nodes, forming the Mix-Enhanced Graph. Finally, we model the relations within the scene using the Graph Encoder and employ a dual-branch diffusion model to generate the corresponding layout and shape, synthesizing the 3D indoor scene.

\subsection{Mixed-Modality Graph}
Generating fine-grained scenes using only text information is insufficient, as it cannot precisely control the geometry of generated objects. At the same time, users' flexible input should be multimodal, allowing for the selective input of text or images based on specific needs, as shown in Fig.~\ref{introduction}.A. However, existing methods \cite{ hu2024scenecraft} do not support this input format. 
Graphs, as a compact and flexible structural representation, enable the effective encoding of diverse attributes within nodes, facilitating the seamless integration of multimodal information. Furthermore, users' textual descriptions often lack information about the relationships between all objects. While methods such as EchoScene\cite{zhai2024echoscene} and CommoScenes \cite{zhai2024commonscenes} utilize graphs to generate scenes, they impose strict relation constraints, making them less user-friendly. A graph structure that mimics natural language should feature sparse edge relations. 
To address these issues, we propose the Mixed-Modality Graph, a novel graph where nodes can contain both textual and visual modalities, and edges are selectable.

A Mixed-Modality Graph $\mathcal{G}^m = \{\mathcal{V}^m, \mathcal{R}^m\}$ contains nodes and their relations:
\begin{equation}
\mathcal{V}^m = \{v_i^m \ | \ i \in {1, \ldots, N}\},
\end{equation}
\begin{equation}
\mathcal{R}^m = \{r_{i \to j}^m \ | \ i, j \in \{1, \ldots, N\}, i \neq j\}.
\end{equation}
Each node $v_i^m = \{[o_i^m, i_i^m] \ | \ [o_i^m] \ | \ [i_i^m]\}$ represents an object with text category information $[o_i^m]$, image information $[i_i^m]$ or both text category and image information $[o_i^m, i_i^m]$, as shown in Fig.~\ref{introduction}.C.
Despite the fact that MMG is generally easier to obtain than 3D spatial layouts \cite{schult2024controlroom3d}, we have also devised a text prompt to query the vision language model such as GPT-4V \cite{achiam2023gpt}, enabling the parsing of MMG from unstructured text and image inputs, as illustrated in Fig.~\ref{introduction}.B. 

Assuming $v_i^m = [o_i^m, i_i^m] $, we utilize embedding layers to encode the category information $o_i^m$ and the relational information of edge $r_{i \to j}^m$, transforming them into $c_i^m$ and $e_{i \to j}^m$, respectively.
To enrich high-level semantic features while simultaneously encoding image information, we leverage the pre-trained and frozen visual language model CLIP \cite{radford2021learning}, using its text encoder to transform $o_i^m$ into $t_i^m$ and its image encoder to convert $i_i^m$ into $u_i^m$.
To ensure consistent processing within MMGDreamer, we apply zero-padding at the feature level for missing node modality information or edge relationships between two nodes.
As illustrated in Fig.~\ref{pipeline}.A, a LMMG $\mathcal{G}_l^m$ can be uniformly represented as:
\begin{equation}
\mathcal{F}_{\mathcal{V}^m} = \left\{ f_i^{v} = [c_i^m, t_i^m, u_i^m] \mid i \in \{1, \ldots, N \} \right\},
\end{equation}
\begin{equation}
\mathcal{F}_{\mathcal{R}^m} = \left\{ f_{i \to j}^{e} = [e_{i \to j}^m] \mid i, j \in \{1, \ldots, N \} , i \neq j\ \right\},
\end{equation}
where $\mathcal{F}_{\mathcal{V}^m}$ represents the set of node features, and $\mathcal{F}_{\mathcal{R}^m}$ represents the set of edge features.

\subsection{Visual Enhancement Module}
Incorporating visual features within graph nodes enhances the generation of object geometry. However, in the LMMG, some nodes only contain textual information. We introduce a visual enhancement module to bolster the ability to generate object shapes. 
This module employs an architecture similar to VQ-VAE, comprising an encoder $E$, a decoder $D$, and a codebook $\mathcal{C}$, to effectively construct visual features from the textual features of nodes within the LMMG. 
The encoder $ E $ processes the textual features $ t_i^m $ into latent vectors $h_i^m=E(t_i^m) $. 
These latent vectors are then quantized using the codebook $\mathcal{C}$, which contains a set of embedding vectors $\{e_k\}_{k=1}^{K}$. The quantization process selects the $n$ nearest embedding vectors from the codebook:
\begin{equation}
\hat{h}_i^m = \{ e_{k_j} \mid k_j \in \arg\min_{e_{k_l} \in \mathcal{C}} \sum_{l=1}^n \| h_i^m - e_{k_l} \|^2 \}.
\end{equation}
The quantized latent vectors $\hat{h}_i^m$ are subsequently processed by the decoder $D$ to generate visual features $\hat{u}_i^m = D(\hat{h}_i^m)$. 
The training objective for the visual enhancement module is to maximize the evidence lower bound (ELBO) for the likelihood of the data:
\begin{equation}
\mathcal{L}_r = \mathbb{E}_{h \sim q_E(h|t)} \left[ \log p_D(u | h) - \beta D_{\text{KL}}(q_E(h|t) \| p(h)) \right],
\end{equation}
where $q_E(h|t)$ denotes the latent vector distribution given the textual features, $p_D(u|h)$ is the likelihood of the visual features given the latent vectors, and $D_{\text{KL}}$ denotes the Kullback-Leibler divergence. 
The prior \( p(h) \) is typically a Gaussian distribution, and \( \beta \) is a weighting factor.
To address the non-differentiable nature of the quantization process, the Gumbel-Softmax relaxation \cite{jang2016categorical} technique is applied to optimize the ELBO. 
Utilizing this VQ-VAE-based framework, the visual enhancement module produces a Visual-Enhanced Graph $\mathcal{G}_I^m$, enhancing the capability of the LMMG to generate accurate and detailed object geometry for scene generation tasks.

\subsection{Relation Predictor}
Relations are crucial in indoor scene generation, as they impact layout configuration. To address the challenge of missing relationships among nodes in the LMMG, we develop a Relation Predictor that infers these connections, enabling the generation of more reasonable layouts.
The Relation Predictor takes triples of latent representations \((f_i^v, f_{i \to j}^e, f_j^v)\) as input.
In cases where relationships are missing, \(f_{i \to j}^e\) is filled with zeros to ensure consistency in the feature space.
The module comprises a GCN layer followed by a series of MLP layers. The GCN layer processes the input triples to capture the relational context between nodes, while the MLP layers further refine the edge predictions.
The Relation Predictor is trained using a cross-entropy loss, defined as: 
\begin{equation}
    \mathcal{L}_e = -\frac{1}{N}\sum_{i=1}^{N}\sum_{c=1}^{C} y_{ic}\log(\hat{y}_{ic})
\end{equation}
where \(N\) is the number of node pairs, \(C\) is the number of edge classes, \(y_{ic}\) is the one-hot encoded true label, and \(\hat{y}_{ic}\) is the predicted probability.
The Relation Predictor refines the graph $\mathcal{G}_I^m$ into a Mixed-Enhanced Graph $\mathcal{G}_E^m$, by predicting and integrating missing node relationships to improve overall layout coherence.

\begin{table*}[t!]
\centering
\setlength{\tabcolsep}{1mm}
    \begin{tabular}{l  c  c c c | c c c | c c c }
     \toprule 
        \multirow{2}{*}{Method} & Shape & \multicolumn{3}{c}{Bedroom} & \multicolumn{3}{c}{Living room} & \multicolumn{3}{c}{Dining room}
        \\ & Representation & FID & $\text{FID}_{\text{CLIP}}$ & KID & FID & $\text{FID}_{\text{CLIP}}$ & KID & FID & $\text{FID}_{\text{CLIP}}$ & KID \\
    \midrule 
    CommonScenes~\cite{zhai2024commonscenes} & rel2shape & 57.68 & \phantom{0}4.86 & \phantom{0}6.59 & 80.99 & \phantom{0}7.05 & \phantom{0}6.39 & 65.71 & \phantom{0}7.04 & \phantom{0}5.47\\
    EchoScene \cite{zhai2024echoscene} & echo2shape &  {48.85} & \phantom{0}{4.26} & \phantom{0}{1.77} &  {75.95} & \phantom{0}{6.73} & \phantom{0}{0.60} & {62.85} & \phantom{0}{6.28} & \phantom{0}{1.72}\\
    \midrule 
    \textbf{MMGDreamer (MM+R)} & echo2shape &  \textbf{45.75}  & \phantom{0}\textbf{3.84} & \phantom{0}\textbf{1.72} &  \textbf{68.94}  & \phantom{0}\textbf{6.19} & \phantom{0}\textbf{0.40} &  \textbf{55.17}  & \phantom{0}\textbf{5.86} & \phantom{0}\textbf{0.05}\\
    \bottomrule
    \end{tabular}
\caption{\textbf{Scene generation realism} is quantified by comparing generated top-down renderings with real scene renderings at a resolution of $256^{2}$ pixels, using FID, $\text{FID}_{\text{CLIP}}$, and KID $(\times 0.001)$, following the methodology in~\cite{zhai2024commonscenes} (lower is better). 
\textbf{MM} denotes nodes using mixed-modality representations. \textbf{R} represents the relationships of nodes. 
}
\label{tab:fidkid}
\end{table*}

\begin{table*}[t]
    \centering
    \setlength{\tabcolsep}{1mm}
    \begin{tabular}{l c|cccccccccc}
    \toprule
    Method & Metric & Bed & N.stand & Ward. & Chair & Table & Cabinet & Lamp & Shelf & Sofa & TV stand\\
    \midrule 
        Graph-to-3D~\cite{dhamo2021graph} & \multirow{4}{*}{MMD} & 1.56 & 3.91 & 1.66 & 2.68 & 5.77 & 3.67 & 6.53 & 6.66 & 1.30 & 1.08 \\
        CommonScenes~\cite{zhai2024commonscenes}& & {0.49} & {0.92} & {0.54} & 0.99 & {1.91} & {0.96} & {1.50} & {2.73} & {0.57} & {0.29} \\
        EchoScene~\cite{zhai2024echoscene} & & 0.37 & 0.75  & 0.39 & 0.62 & 1.47 & 0.83 & 0.66 & 2.52 & 0.48 & 0.35 \\
        \textbf{MMGDreamer (I+R)} & & \textbf{0.22} & \textbf{0.41} & \textbf{0.24}  & \textbf{0.35} & \textbf{0.55} & \textbf{0.71} & \textbf{0.34} & \textbf{1.58} & \textbf{0.43} & \textbf{0.24}  \\
    \midrule 
        Graph-to-3D~\cite{dhamo2021graph} & \multirow{4}{*}{COV} & 4.32 & 1.42 & 5.04 & 6.90 & 6.03 & 3.45 & 2.59 & 13.33 & 0.86 & 1.86 \\
        CommonScenes~\cite{zhai2024commonscenes}& & {24.07} & {24.17} & {26.62} & {26.72} & {40.52} & {28.45} & {36.21} & {40.00} & {28.45} & {33.62} \\
        EchoScene~\cite{zhai2024echoscene} & & 39.51 & 25.59 & 37.07 & 17.25 & 35.05 & 43.21 & 33.33 & 50.00 & 41.94 & 40.70\\
        \textbf{MMGDreamer (I+R)} & & \textbf{42.59} & \textbf{30.81}  & \textbf{44.44} & \textbf{19.95} & \textbf{44.12} & \textbf{49.38} & \textbf{40.56} & \textbf{70.00} & \textbf{47.31} & \textbf{45.35} \\
    \midrule 
        Graph-to-3D~\cite{dhamo2021graph} & \multirow{4}{*}{1-NNA} & 98.15 & 99.76 & 98.20 & 97.84 & 98.28 & 98.71 & 99.14 & 93.33 & 99.14 & 99.57 \\
        CommonScenes~\cite{zhai2024commonscenes}& & {85.49} & {95.26} & {88.13} & {86.21} & {75.00} & {80.17} & {71.55} & {66.67} & {85.34} & {78.88} \\
        EchoScene~\cite{zhai2024echoscene} & & 72.84 & 91.00 & 81.90 & 92.67 & 75.74 & {69.14} & 78.90 & \textbf{35.00} & 69.35 & 78.49\\
        \textbf{MMGDreamer (I+R)} & & \textbf{69.44} &  \textbf{90.52} & \textbf{74.81} & \textbf{89.56} & \textbf{68.85} & \textbf{68.35} & \textbf{72.38} & {30.00} & \textbf{62.37} & \textbf{73.26} \\
    \bottomrule
    \end{tabular}
    \caption{\textbf{Object-level generation performance.} We present MMD ($\times0.01$, $\downarrow$), COV($\%, \uparrow$), and 1-NNA($\%, \downarrow$) metrics to assess the quality and diversity of the generated shapes. \textbf{I} denotes nodes using image representations. \textbf{R} represents node relationships.}
    \label{tab:objfd}
\end{table*}

\begin{figure*}[t]
\centering
\includegraphics[width=1\textwidth]{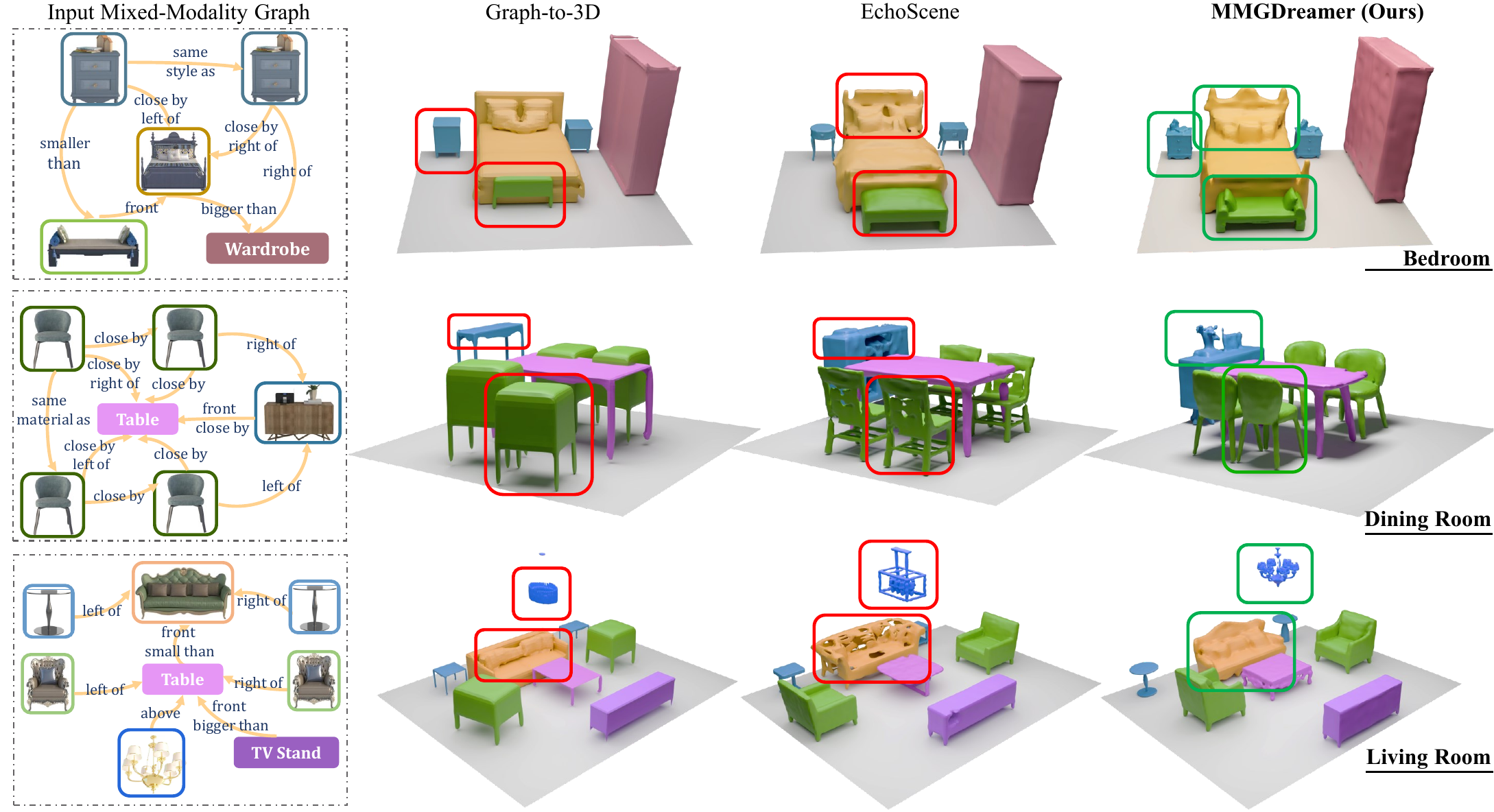} 
\caption{\textbf{Qualitative comparison} with other methods. The first column shows the input mixed-modality graph, which visualizes only the most critical edges in the scene. Red rectangles denote areas of inconsistency in the generated scenes, while green rectangles signify regions of consistent generation.}
\label{demo}
\end{figure*}

\subsection{Shape and Layout Branch}
We employ a dual-branch diffusion model to generate object shapes and scene layouts. 
To facilitate effective information exchange and relationship modeling between nodes during each denoising process, as depicted in Fig.~\ref{pipeline}.C.1, we employ a triplet-GCN structured module that integrates the echo mechanism \cite{zhai2024echoscene} as a Graph Encoder $E_g$.

\subsubsection{Shape Branch.}
For the shape branch, as shown in Fig.~\ref{pipeline}.C.2, we use Truncated Signed Distance Field \cite{curless1996volumetric} as shape representations and employ a pretrained and frozen VQ-VAE to encode them into latent representations $z_i^0$ and decode them back.
At each denoising step \( t \in \{1, 2, \dots, T\} \), $E_g$ is applied to process the latent codes \( z_i^t \) and the latent graph \( \mathcal{G}_t^l \) (which originates from \( \mathcal{G}_E^m \)), yielding updated representations \( \tilde{z}_i^t \) and \( \tilde{\mathcal{G}_t^l} \).
The updated nodes of  \( \tilde{\mathcal{G}_t^l} \), denoted as \( \tilde{\mathcal{V}_t} = \{ \tilde{v}_i^t \} \), are used as conditions for denoiser \( \epsilon_{\theta} \) (3D-UNet). 
The training objective is to minimize the deviation between the true noise \( \epsilon \) and the predicted noise \( \epsilon_{\theta} (\tilde{z}_i^t, t, \tilde{v}_i^t) \). The loss function is defined as:
\begin{equation}
\mathcal{L}_s = \mathbb{E}_{\tilde{z}_i^t, \epsilon \sim \mathcal{N}(0,1), t} \left\| \epsilon - \epsilon_{\theta} (\tilde{z}_i^t, t, \tilde{v}_i^t) \right\|^2.
\end{equation}

\subsubsection{Layout Branch.}
We utilize object bounding boxes to represent the layout of the scene.
Each bounding box \( l_i^0 \) is characterized by its location \( \mathbf{t}_i^0 \in \mathbb{R}^3 \), size \( \mathbf{s}_i^0 \in \mathbb{R}^3 \), and rotation angle \( \varphi_i^0 \). Specifically, the rotation angle \( \varphi_i^0 \) is parameterized by \( [\cos( \varphi_i^0), \sin( \varphi_i^0)]^\top \). 
To ensure proper scale and numerical stability during training, \( \mathbf{t}_i^0 \) and \( \mathbf{s}_i^0 \) are normalized.
As illustrated in Fig.~\ref{pipeline}.C, the layout branch utilizes $E_g$ for relationship modeling. This results in updated latent layout representations \( \tilde{l}_i^t \) and refined graph node embeddings  \( \tilde{\mathcal{N}}_t = \{ \tilde{n}_i^t \} \).
Conditioned on the updated node embeddings, a 1D-UNet is utilized as the denoiser $\psi_{\theta}$ for the denoising process.
The corresponding loss function is formulated as:
\begin{equation}
\mathcal{L}_l = \mathbb{E}_{\tilde{l}_i^t, \psi \sim \mathcal{N}(0,1), t} \left\| \psi - \psi_{\theta} (\tilde{l}_i^t, t, \tilde{n}_i^t) \right\|^2.
\end{equation}
The overall training objective for the layout and shape branches is expressed as:
\begin{equation}
    \mathcal{L}_o = \alpha_1 \mathcal{L}_{l} + \alpha_2 \mathcal{L}_{s},
\end{equation}
where $\alpha_1$ and $\alpha_2$ are weighting factors.
\subsection{Training and Inference Strategy}
The training process is divided into two stages. In the first stage, the visual enhancement module is trained with the loss function $\mathcal{L}_r$, which utilizes textual information from nodes to construct the corresponding visual features. The Relation Predictor is trained with $\mathcal{L}_e$ using triplet representations of the graph. 
In the second stage, the LMMG serves as the input, and the loss function $\mathcal{L}_o$ is employed to jointly optimize the graph encoder with the layout and shape branches, as depicted in Fig.~\ref{pipeline}.A and C.
During inference, as shown in Fig.~\ref{pipeline}, the LMMG is processed through modules B and C to generate the indoor scene. Please see the Supplementary Material for further details.

\section{Experiments}

\subsection{Experimental Settings}

\subsubsection{Evaluation Dataset.}
We validate our approach using the SG-FRONT dataset \cite{zhai2024commonscenes}, which provides comprehensive scene-graph annotations for indoor scenes. 
This dataset includes 45K object instances and 15 types of relationships within bedrooms, dining rooms, and living rooms.
Nodes in the scene graphs represent object categories, while edges indicate relationships between the nodes.
In our experiments, we extracted corresponding images from the 3D-FUTURE dataset \cite{fu20213d_future} based on node IDs to construct a Full-Modality Graph (node contains text and image). We then applied a random mask to mask the text, images, and relationships between nodes in the Full-Modality Graph, producing the Mixed-Modality Graph.
\subsubsection{Evaluation Metrics.}
We evaluate the scene-level and object-level fidelity of the synthesized 3D scenes. 
Scene-level fidelity is quantified using Fŕchet Inception Distance (FID) \cite{heusel2017gans} and Kernel Inception Distance (KID) \cite{binkowski2018demystifying}, which measure the similarity between generated top-down renderings and real scenes renderings. 
For object-level fidelity, we assess the quality of generated object geometry using Minimum Matching Distance (MMD), Coverage (COV), and 1-Nearest Neighbor Accuracy (1-NNA) \cite{yang2019pointflow}, all derived from Chamfer Distance (CD) \cite{fan2017point}.

\subsubsection{Baselines.}
We compare our approach with three state-of-the-art scene synthesis methods: 
1) \textbf{Graph-to-3D} \cite{dhamo2021graph}, which generates 3D scenes directly from scene graphs using a GCN-based VAE; 
2) \textbf{CommonScenes} \cite{zhai2024commonscenes}, which converts scene graphs into controllable 3D scenes through a dual-branch framework with a VAE and LDM; 
3) \textbf{EchoScene} \cite{zhai2024echoscene}, which employs a dual-branch diffusion model with an information echo mechanism for generating globally coherent 3D scenes from scene graphs.

\subsubsection{Implementation Details.}
All experiments are performed on a single NVIDIA A100 GPU with 80 GB memory.
We train our models using the AdamW optimizer, initializing the learning rate at $1 \times 10^{-4}$ and utilizing a batch size of 128. The weighting factors for our loss components, $\alpha_1$ and $\alpha_2$, are consistently set to 1.0.

\subsection{Scene Generation}

\begin{table}[t!]
\centering
    \begin{tabular}{llccc}
    \toprule
         VEM & RP & FID & KID & mSG \\
     \midrule
           & & 45.10& 3.74 & 0.63 \\
          \phantom{0}\checkmark & & 43.50& 3.27 & 0.63\\
           & \checkmark & 44.07 & 3.43& 0.83 \\
    \midrule 
         \phantom{0}\checkmark & \checkmark&\textbf{ 41.84}& \textbf{2.55}& \textbf{0.86}\\
     \bottomrule
    \end{tabular}
\caption{\textbf{Ablation studies.} Visual Enhancement Module and Relation Predictor are abbreviated as VEM and RP, respectively. The best results are highlighted in \textbf{bold}.}
\label{tab:ablation}

\end{table}

\subsubsection{Quantitative Comparison.}
We evaluate the realism of generated scenes using FID, FID$_{\text{CLIP}}$, and KID scores, as detailed in Tab. \ref{tab:fidkid}. 
MMGDreamer consistently outperforms the previous state-of-the-art method Echoscene, across all metrics when scene graph nodes are represented with mixed-modality. 
Specifically, in living room generation, MMGDreamer (MM+R) achieves a remarkable improvement, reducing FID by 9\%, FID$_{\text{CLIP}}$ by 8\%, and KID by 33\%, highlighting its superior ability to control object geometry while enhancing overall scene realism.

\subsubsection{Qualitative Comparison.}
We present the generated results for different methods across various room types in Fig.~\ref{demo}. In comparison across different room types, our method MMGDreamer consistently demonstrates superior geometry control and visual fidelity in every scenario. 
For instance, in the bedroom, MMGDreamer accurately generates the bed and nightstands with higher geometric consistency, while other methods like Graph-to-3D and EchoScene display noticeable distortions and inconsistencies.
In the dining room, both Graph-to-3D and EchoScene display significant deficiencies, particularly with chair backrests and the sideboard. In contrast, our method, MMGDreamer, not only preserves the correct geometry of these elements but also successfully generates the intricate details of objects placed on the sideboard.
For the complex living room scene, MMGDreamer accurately generates the sofa, coffee table, and lamp, maintaining a coherent spatial layout and ensuring a high degree of consistency between the generated objects and the input images. By contrast, other methods exhibit geometry errors in several pieces of furniture, such as the lamp and chair. 
Notably, the sofa generated by EchoScene contains numerous visible holes, significantly deviating from the actual object geometry.

\subsection{Object Generation}
We extend our analysis to the object level fidelity, following PointFlow \cite{yang2019pointflow}, by reporting the MMD (×0.01), COV (\%), and 1-nearest neighbor accuracy (1-NNA, \%) metrics to assess per-object generation.
As presented in Tab. \ref{tab:objfd}, our method consistently surpasses the previous state-of-the-art across all object categories in both MMD and COV metrics.
This result highlights MMGDreamer’s geometric control capabilities, ensuring the precise generation of object geometry across various categories.
The 1-NNA measures the distributional similarity between the generated objects and the ground truth, with values near 50\% indicating better capturing of the shape distribution.
Across most object categories, our method consistently outperforms EchoScene in terms of distributional similarity. 
Overall, MMGDreamer demonstrates superior geometric control, resulting in more consistent object-level generation compared to previous approaches.

\subsection{Ablation Study}
We utilize scene-level fidelity (FID and KID) and mean scene graph consistency (mSG) to quantitatively evaluate the effectiveness of different modules within MMGDreamer, as presented in Tab. \ref{tab:ablation}. 
We observe that the configuration with VEM (second row) shows a significant decrease in FID and KID compared to the baseline (first row), indicating that VEM enhances the fidelity of scene generation. Additionally, when the RP module is introduced (third row), there is a notable improvement in mSG, demonstrating that RP effectively predicts relationships between objects, resulting in more coherent scene layouts. It is evident that including both VEM and RP achieves the best performance across all metrics, highlighting the complementary benefits of these modules in producing high-quality scene generation.

\section{Conclusion}
We present MMGDreamer, a dual-branch diffusion model for geometry-controllable 3D indoor scene generation, leveraging a novel Mixed-Modality Graph that integrates both textual and visual modalities. Our approach, enhanced by a Visual Enhancement Module and a Relation Predictor, provides precise control over object geometry and ensures coherent scene layouts. 
Extensive experiments demonstrate that MMGDreamer significantly outperforms existing methods, achieving state-of-the-art results in scene fidelity and object geometry controllability.

\section{Acknowledgments}
The authors would like to thank the anonymous reviewers for their comments. This work was supported by the National Key R\&D Program of China under Grant 2023YFB2703800 and the Beijing Natural Science Foundation under Funding No. IS23055. The contact author is Zhen Xiao and Chao Zhang.

\bibliography{aaai25}
\appendix

\renewcommand{\thetable}{\arabic{table}}
\renewcommand{\thefigure}{\arabic{figure}}
\renewcommand{\thesection}{A\arabic{section}}
\renewcommand{\theequation}{\arabic{equation}}

\setcounter{table}{0}
\setcounter{section}{0}
\setcounter{equation}{0}
\setcounter{figure}{0}

\twocolumn[{
\centering
\section*{\LARGE \centering Supplementary Material of MMGDreamer}
 \vspace{30pt}
 }]

In this Supplementary Material, we report the following:
\begin{itemize}
\item Section~\ref{sec:Additional Results}: Additional Results.
\item Section~\ref{sec:More Qualitative Results}: Additional Qualitative.
\item Section~\ref{sec:Additional Experimental Details}: Additional Experimental Details.
\item Section~\ref{sec:Limitations and Future Work}: Limitations and Future Work.
\end{itemize}

\section{Additional Results}
\label{sec:Additional Results}


\subsection{Scene Graph Manipulation}

\begin{table*}[t!]
    \caption{\textbf{Scene graph consistency} (higher is better). \textbf{MM} represents nodes using mixed-modality representations. \textbf{R} denotes the relationships of nodes. Top to bottom: Relationship change mode, Node addition mode, and Generation only.}
    \centering
    \scalebox{0.86}{
    \begin{tabular}{l c |c| cccc cc}
    \toprule
    \multirow{2}{*}{Method} & \multirow{2}{*}{\makecell{Shape \\ Representation}} & \multirow{2}{*}{\makecell{Mode}} & \multirow{2}{*}{\makecell{left/\\right}} & \multirow{2}{*}{\makecell{front/\\behind}} & \multirow{2}{*}{\makecell{smaller/\\larger}} & \multirow{2}{*}{\makecell{taller/\\shorter}} & \multirow{2}{*}{\makecell{close by}} & \multirow{2}{*}{\makecell{symmetrical}} \\
    & & & & & & & &\\
        
    \midrule 
        \cmidrule{1-2} \cmidrule{4-9}
        Graph-to-3D~\cite{dhamo2021graph} &  DeepSDF~\cite{park2019deepsdf} & \multirow{4}{*}{Change} & 0.91 & 0.92 & 0.86 & 0.89 & 0.69 & 0.46 \\
        CommonScenes~\cite{zhai2024commonscenes} & rel2shape & & 0.91 & 0.92 & 0.86 & 0.91 & 0.69 & {0.59} \\
         EchoScene~\cite{zhai2024echoscene}& echo2shape & & {0.94} & \best{0.96} & {0.92} & {0.93} & \best{0.74} & 0.50\\
         \textbf{MMGDreamer (MM+R)} & echo2shape && \best{0.95}& \best{0.96} & \best{0.93} & \best{0.93} & 0.71 & \best{0.53}  \\
        \midrule
        \cmidrule{1-2} \cmidrule{4-9}
        Graph-to-3D~\cite{dhamo2021graph} &  DeepSDF~\cite{park2019deepsdf} & \multirow{4}{*}{Addition} &  0.94 & 0.95 & 0.91 & 0.93 & 0.63 & 0.47\\
        CommonScenes~\cite{zhai2024commonscenes} & rel2shape & & 0.95 & 0.95 & 0.91 & 0.95 & 0.70 & \best{0.61}\\
         EchoScene~\cite{zhai2024echoscene} & echo2shape & & \best{0.98} & {0.99} & {0.96} & {0.96} & {0.76} & 0.49\\
         \textbf{MMGDreamer (MM+R)} & echo2shape & & \best{0.98} & \best{1.00} & \best{0.97} & \best{0.97} & \best{0.80} & \best{0.61} \\
    \midrule
    \cmidrule{1-2} \cmidrule{4-9}
        Graph-to-3D~\cite{dhamo2021graph} & DeepSDF~\cite{park2019deepsdf} & \multirow{4}{*}{None} & \best{0.98} & 0.99 & 0.97 & 0.95 & 0.74 & 0.57\\
        CommonScenes \cite{zhai2024commonscenes} & rel2shape & & \best{0.98} & \best{1.00} & 0.97 & 0.95 & \best{0.77} & {0.60} \\
        EchoScene~\cite{zhai2024echoscene} & echo2shape & & \best{0.98} & 0.99 & {0.96} & \best{0.96} & 0.74 & 0.55\\
        \textbf{MMGDreamer (MM+R)} & echo2shape & & \best{0.98} & {0.99} & \best{0.97} &\best{0.96}  &{0.76}  & \best{0.62} \\
    \bottomrule
    \end{tabular}}
    \label{tab:sgconst}
\end{table*}

\begin{table*}[t!]
\caption{\textbf{Inter-object consistency.} The object shapes corresponding to the \textit{same as} relationship within a scene demonstrate a high degree of consistency, as reflected by the low CD values (scaled by $\times 0.001$). \textbf{I} represents nodes using image representations. \textbf{R} denotes the relationships of nodes.}
\centering
\scalebox{1}{
\begin{tabular}{l ccc | ccc | ccc}
    \toprule
         \multirowcell{2}{Method} & \multicolumn{3}{c}{Bedroom} & \multicolumn{3}{c}{Living room} & \multicolumn{3}{c}{Dining room}\\
         \cmidrule{2-10}
         & Wardrobe & Nightstand &Lamp & Chair & Table &Lamp& Chair & Table & Sofa\\
     \midrule
         CommonScenes \cite{zhai2024commonscenes} & 0.61&2.69& - &6.64&11.75 & - & 1.96 &  9.04 & - \\
         EchoScene \cite{zhai2024echoscene} &0.14 & 1.68& 30.07&0.99& 3.02 & 10.06&1.75  & 1.26 &3.47 \\
         \textbf{MMGDreamer~(I+R)}  & \best{0.11} & \best{1.33}& \best{2.29}& \best{0.16}& \best{2.44} & \best{0.18} &\best{0.23} &\best{0.83}& \best{0.18}  \\
     \bottomrule
    \end{tabular}
    }
\label{tab:consistency}
\end{table*}

We follow EchoScene \cite{zhai2024echoscene} in manipulating scene graphs by either adding a node with relevant edges or altering the relationships between existing nodes, the results presented in Tab. \ref{tab:sgconst}.
Our method, MMGDreamer (MM+R), consistently outperforms other approaches in most categories, particularly excelling in the "left/right," "smaller/larger," and "taller/shorter" relationships, where it achieves the highest scores. For example, in the 'left/right' category for relationship change mode, MMGDreamer (0.95) outperforms both Graph-to-3D (0.91) and CommonScenes (0.91), demonstrating its ability to maintain consistency between the generated scene's spatial relationships and the input graph structure. Notably, in the Node addition mode, MMGDreamer achieves a perfect score 1.00 in the "front/behind" category, indicating its superior capability in preserving spatial relationships during scene graph manipulations.
Across all manipulation modes, MMGDreamer (MM+R) demonstrates a clear superiority in the symmetrical metric compared to other methods. This consistent performance underscores the advantage of incorporating visual information into the mixed-modality graph, which enables more precise geometry control and leads to the generation of scenes with objects that exhibit enhanced symmetry.

\subsection{Inter-object Consistency}
To evaluate inter-object consistency, we measure the Chamfer Distance (CD) values for object shapes that share the \textit{same as} relationship within a scene. Low CD values indicate higher consistency in object shapes. 
As shown in Tab.~\ref{tab:consistency}, MMGDreamer (I+R) consistently achieves lower CD values across various objects and room types, demonstrating stronger geometry control and significantly higher inter-object consistency compared to CommonScenes and EchoScene.
For example, in the Bedroom, MMGDreamer reduces the CD for the Nightstand to 1.33, which is 1.36 lower than CommonScenes and 0.35 lower than EchoScene. Even for the Lamp, where EchoScene performs poorly with a CD of 30.07, MMGDreamer (I+R) shows much better consistency with a CD of 2.29, representing an improvement of 27.78, demonstrating MMGDreamer's ability to maintain object shape consistency even in more challenging object types.
In the Living room, MMGDreamer achieves a CD of 0.16 for the Chair, outperforming EchoScene by 0.83. For the Table, MMGDreamer (I+R) yields a CD of 2.44, which is 8.31 lower than CommonScenes and 0.58 lower than EchoScene.
In the Dining room, MMGDreamer maintains a low CD of 0.83 for the Table, a notable improvement of 8.21 over CommonScenes and 0.43 over EchoScene.
Overall, MMGDreamer (I+R) demonstrates superior control over object geometry, as evidenced by consistently lower CD values across Bedroom, Living room, and Dining room scenes.
\section{Additional Qualitative}
\label{sec:More Qualitative Results}

\subsection{Qualitative Results On Scene Generation}
\begin{figure*}[t]
\centering
\includegraphics[width=1\textwidth]{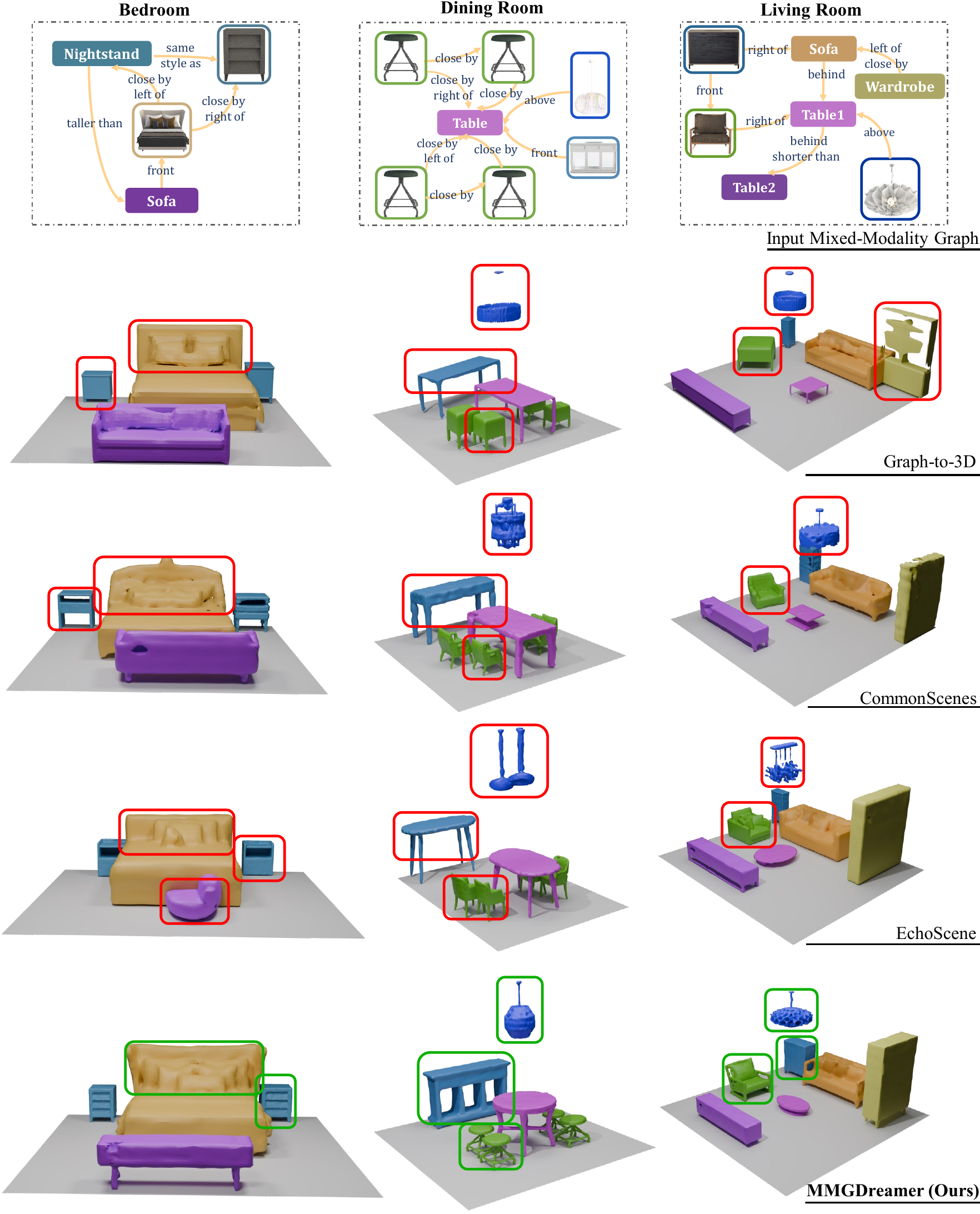} 
\caption{\textbf{More qualitative comparison on scene generation.} The first row shows the input mixed-modality graph, which visualizes only the most critical edges in the scene. Red rectangles denote areas of inconsistency in the generated scenes, while green rectangles signify regions of consistent generation.}
\label{sup_scene_demo}
\end{figure*}

We present a qualitative comparison of scene generation results across different methods, as shown in Fig.~\ref{sup_scene_demo}.
MMGDreamer consistently excels in maintaining object geometry and spatial relationships, resulting in more detailed and realistic scenes compared to other methods.
For example, in the bedroom scene, MMGDreamer successfully maintains the precise geometric alignment between the Nightstand and Bed, as indicated by the green rectangles. 
In contrast, Graph-to-3D and CommonScenes exhibit issues with the geometry of the Nightstand and Bed, leading to unrealistic shapes. In particular, EchoScene generates a visibly distorted Sofa with incorrect placement, leading to significant inconsistencies in both shape and spatial location.
In the dining room scene, MMGDreamer accurately captures the complex geometry of the Lamp and maintains the correct spatial relationship between the Chair and Table. Other methods, like CommonScenes and EchoScene, struggle to reproduce the Lamp's intricate details, leading to visible distortions, and fail to maintain the correct positioning of the Chair and Table. This highlights MMGDreamer's clear advantage in handling both complex shapes and spatial relationships.
In the complex living room scene, MMGDreamer effectively demonstrates superior geometry controllability by accurately generating the sofa, chair, and lamp, maintaining both precise object shapes and a consistent spatial arrangement that closely aligns with the input graph. In contrast, other methods exhibit significant geometry issues, particularly with the chair and lamp.
\subsection{Qualitative Results On Object Generation}
\begin{figure*}[t]
\centering
\includegraphics[width=0.9\textwidth]{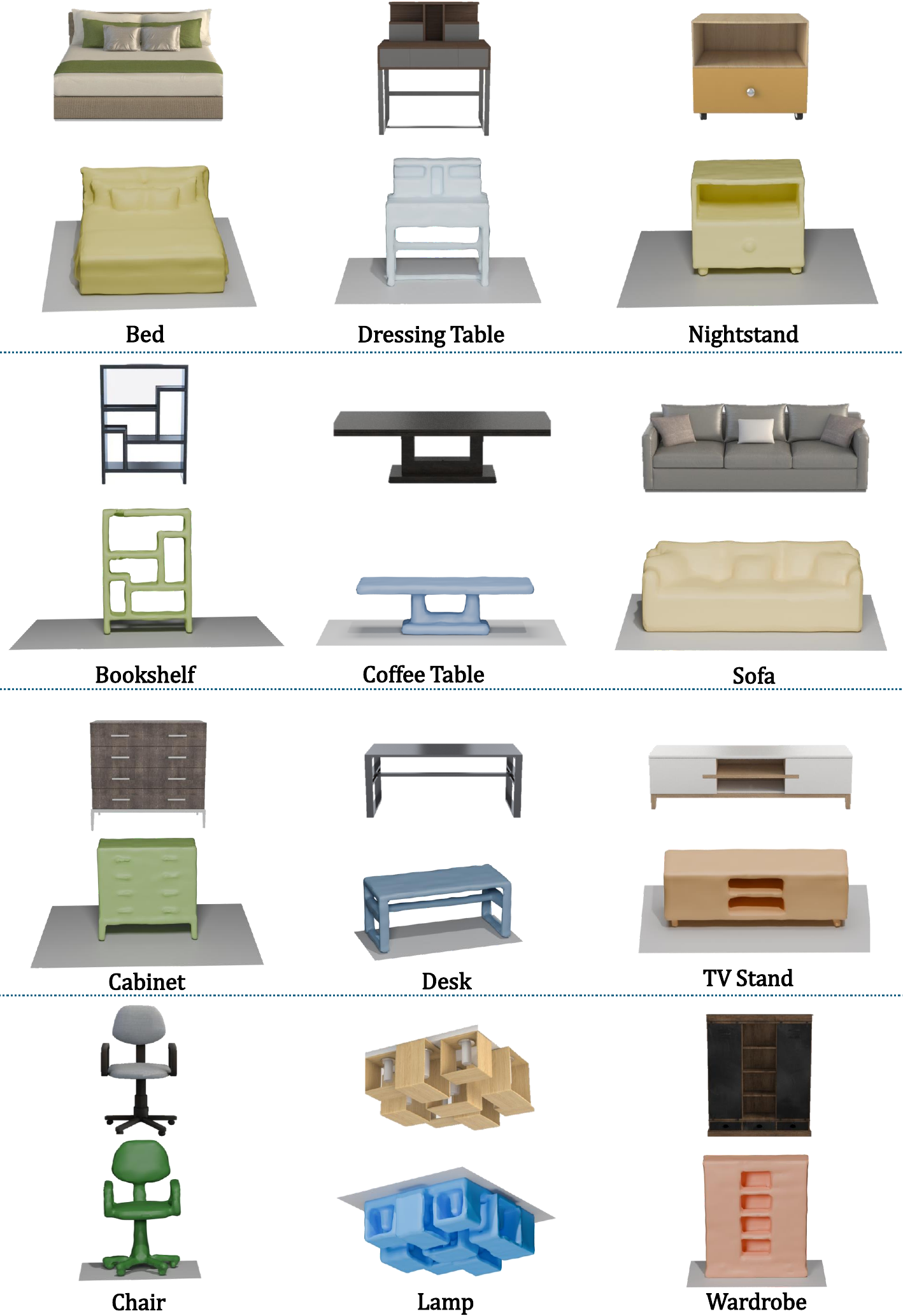} 
\caption{\textbf{Qualitative results on object generation.} The figure is divided into three sections by dashed lines. In each section, the top row shows the input images of various furniture items, the middle row displays the corresponding generated objects in the scenes, and the bottom row provides the object categories.}
\label{sup_obj_demo}
\end{figure*}

We provide a qualitative analysis of the objects generated by MMGDreamer (I+R) in Fig.~\ref{sup_obj_demo}. 
The results demonstrate a high degree of consistency between the generated object shapes and the input images, showcasing the strong geometry controllability of our method.
For example, in generating complex objects like the Chair and Lamp, MMGDreamer successfully produces highly consistent geometries.
The Chair, with its intricate structure and unique shape, is accurately captured in the generated object, maintaining consistency with the input image in both shape and proportions. Similarly, the Lamp's complex geometry and fine details are faithfully reproduced, showcasing our model's high precision in capturing and generating intricate shapes.
Compared to methods specifically designed for object generation, such as One-2-3-45++ \cite{liu2024one}, which require large amounts of training data, MMGDreamer (I+R) achieves impressive object geometry generation results with only a small amount of training data.
This demonstrates the robustness and geometric controllability of our approach, even under data-limited conditions, while still generating high-quality object shapes.

\subsection{Qualitative results on relation-free scene generation}
\begin{figure*}[t]
\centering
\includegraphics[width=1\textwidth]{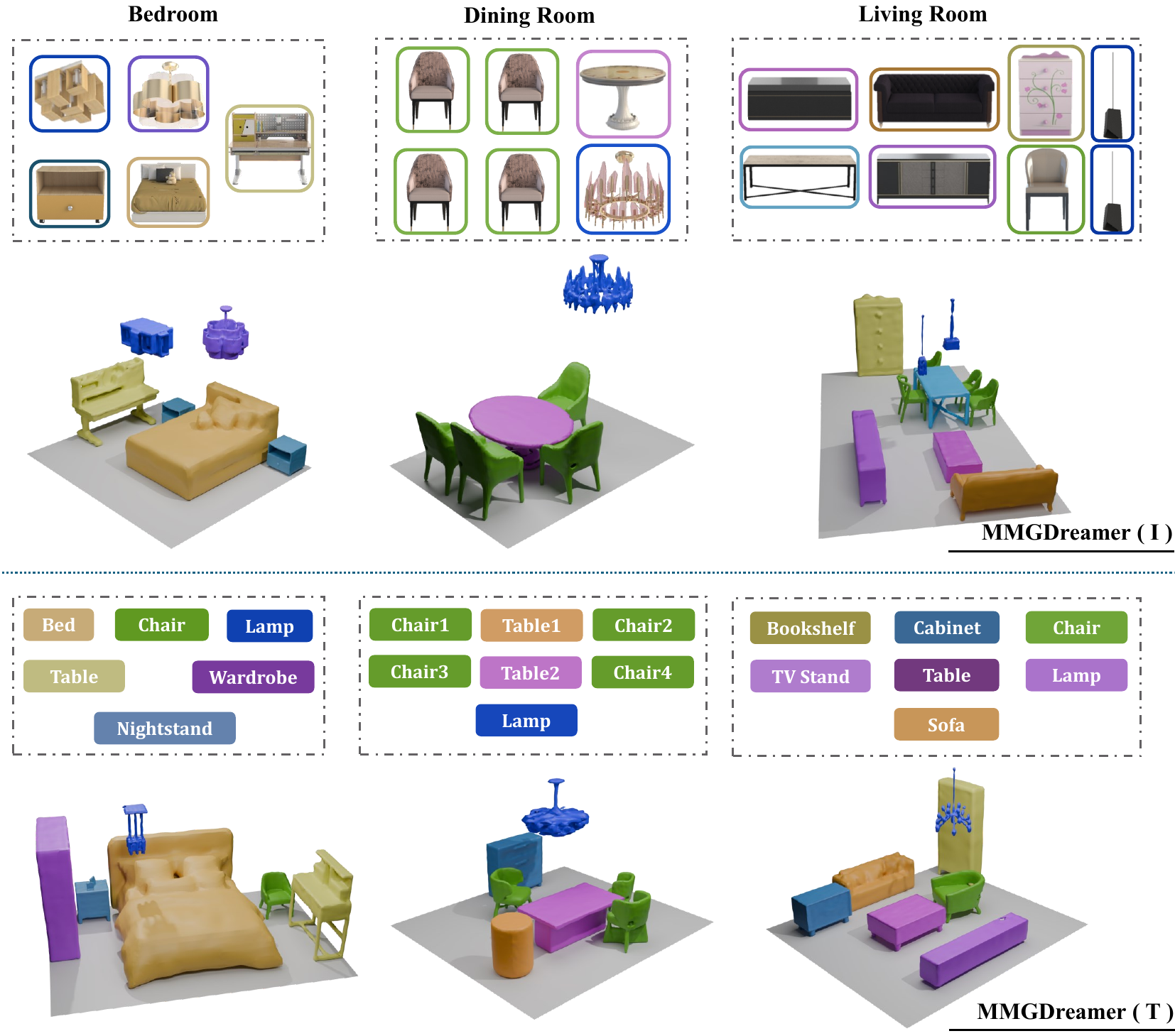} 
\caption{\textbf{Qualitative Results On Relation-Free Scene Generation.} The figure is divided into two sections by dashed lines. In each section, the dashed boxes represent the input mixed-modality graphs, where nodes are depicted either as text or images, without any explicit relationships. Below each input graph, the corresponding generated indoor scenes are displayed.}
\label{sup_relation_free}
\end{figure*}

We demonstrate the generated results of MMGDreamer (I) and MMGDreamer (T) when provided with mixed-modality graphs that lack any explicit object relationships, as shown in Fig.~\ref{sup_relation_free}.
Despite the absence of predefined relationships, our method successfully generates coherent and realistic layouts. This highlights the effectiveness of the Relation Predictor within MMGDreamer, which can infer the spatial relationships between objects, leading to well-organized scene layouts.
For example, in the Bedroom scene generated by MMGDreamer (I), the bed, nightstands, and lamp are not only arranged logically but also exhibit a high degree of fidelity. The objects' geometries in the generated scene closely match the corresponding input images, showcasing MMGDreamer (I)'s ability to maintain geometric consistency and high detail throughout the scene generation process. Similarly, MMGDreamer (T) successfully arranges the objects in the Living Room scene, where the sofa, tables, and chairs are organized into a cohesive layout that reflects real-world spatial arrangements, again without any predefined relationships.
These results demonstrate the robustness of MMGDreamer’s Relation Predictor, which predicts object relationships and generates reasonable layouts under relation-free conditions, ensuring consistent and visually harmonious scene generation.

\section{Additional Experimental Details}
\label{sec:Additional Experimental Details}

\subsection{Baselines}

\subsubsection{Graph-to-3D.}
Graph-to-3D \cite{dhamo2021graph}is an approach that directly generates 3D shapes from a scene graph in an end-to-end manner. Unlike previous methods that rely on retrieving object meshes from synthetic data, Graph-to-3D leverages GCN within a variational autoencoder framework to generate both object shapes and scene layouts. This model allows for flexible scene synthesis and modification, using the scene graph as an interface for semantic control, providing a more robust and direct method for 3D scene generation.
We utilize the DeepSDF \cite{park2019deepsdf} variant of Graph-to-3D for SDF-based shape generation, training twelve category-specific models (excluding "floor") for 1500 epochs using SG-FRONT. The latent codes for each object are optimized and stored, then used to train Graph-to-3D. During inference, the model directly generates 3D shapes and full scenes using the predicted latent codes.

\subsubsection{CommonScenes.}
CommonScenes \cite{zhai2024commonscenes} is a fully generative model that effectively converts scene graphs into controllable 3D scenes that are semantically realistic and conform to commonsense. Unlike previous methods that rely on database retrieval or pre-trained embeddings, CommonScenes uses a dual-branch pipeline to predict scene layouts and generate object shapes while capturing global and local relationships in the scene graph. 
We follow the training procedure outlined in \cite{zhai2024commonscenes} and train the network end-to-end on the SG-FRONT dataset using the AdamW optimizer with an initial learning rate of $1 \times 10^{-4}$ for 1000 epoch.

\subsubsection{EchoScenes.}
EchoScene \cite{zhai2024echoscene} is a generative model designed to create 3D indoor scenes from scene graphs by utilizing a dual-branch diffusion model. It handles the complexities of scene graphs, such as varying node counts and diverse edge combinations, by introducing an information echo scheme. This allows for collaborative information exchange between nodes, ensuring that the generated scenes are both globally coherent and controllable.
Adhering to the training protocol from \cite{zhai2024echoscene}, we trained EchoScene on the SG-FRONT dataset for 2050 epochs.

\subsubsection{Text-to-shape Series.}
This series includes two generative baselines. One is built upon CommonScenes, called CommonLayout+SDFusion, and the other builds upon Echoscene, referred to as EchoLayout+SDFusion. Both methods first generate bounding boxes and then use the text-to-shape method SDFusion \cite{cheng2023sdfusion} to further generate shapes within each bounding box, based on the textual information from the graph nodes.

\subsection{Implementation Details}

\subsubsection{Hardware and Software.}
We demonstrate the hardware and software specifications of our experimental setup, including CPU, GPU, and system configuration, as shown in Tab.~\ref{tab: hardware}. 
In addition, we utilize Blender 4.1 with the CYCLES engine to render high-quality images for our qualitative comparison experiments. In our Blender setup, we configure the Noise Threshold to 0.001, set the maximum samples to 300, and use the RGBA color mode with a color depth of 16. 
We also ensure that these parameters remain consistent across all qualitative comparison experiments to maintain uniformity in the rendering process.

\begin{table}[!t]
\centering
\caption{\textbf{Hardware and software} specifications for experimental setup.}
\scalebox{0.91}{
\begin{tabular}{cc} 
\toprule
\multicolumn{2}{c}{\textbf{System \& Hardware Overview}}  \\ 
\midrule
\multirow{2}{*}{CPU} & Intel(R) Xeon(R) Platinum~             \\
                     & 8375C CPU @ 2.90GHz                    \\
GPU                  & 8 $\times$NVIDIA A100 Tensor Core GPU    \\
Memory               & 10T  DRAM                                 \\
Operating System     & Ubuntu 22.04.4 LTS                     \\
CUDA Version         & 11.3                                   \\
NVIDIA Driver        & ~530.30.02                             \\
ML Framework         & Python 3.8.18  Pytorch 1.11.0          \\

\midrule
\multicolumn{2}{c}{\textbf{GPU Specifications}}  \\ 
\midrule
CUDA Cores& 6912 \\
Memory Capacity& 80GB \\
Memory Bandwidth& 1935GB/s \\
\bottomrule
\end{tabular}
}
\label{tab: hardware}
\end{table}
\subsubsection{Dataset Details.}
Our experiments are conducted on SG-FRONT, a dataset that enhances the 3D-FRONT synthetic dataset by incorporating comprehensive scene graph annotations. These annotations are organized into three key categories: spatial/proximity, support, and style relationships. Spatial relationships dictate object positions (e.g., left/right), size comparisons (e.g., bigger/smaller), and height comparisons (e.g., taller/shorter). Support relationships capture structural dependencies such as proximity and relative placement (e.g., close by, above, standing on). Style relationships reflect similarities in material, shape, and category. 
SG-FRONT contains around 45k samples, covering three types of indoor scenes: bedrooms, dining rooms, and living rooms, with annotations for 15 different relationship types.
We follow the training and testing procedures outlined in EchoScene \cite{zhai2024echoscene} to assess all methods on SG-FRONT. The dataset consists of 4041 bedrooms, 900 dining rooms, and 813 living rooms. For training, we use 3879 bedrooms, 614 dining rooms, and 544 living rooms, while the remaining scenes are reserved for testing.

\subsubsection{ChatGPT Prompt.}
The prompt for Mixed-Modality Graph Generation using GPT-4V is shown in Fig.~\ref{fig:prompt-gpt}. The design of this prompt focuses on enabling GPT-4V to effectively interpret and generate structured scene graphs from both text descriptions and image inputs. By leveraging GPT-4V's multimodal capability, the prompt enables seamless integration of diverse inputs, ensuring that all relationships between objects are captured accurately and consistently within the generated scene graph.

\subsubsection{Batch Size Definition.}
During the training of the dual-branch diffusion model, we follow the approach used in EchoScene \cite{zhai2024echoscene}, where each branch operates with its batch size to accommodate distinct training objectives. 
For the layout branch, we define a scene batch $B_l$, which includes all bounding boxes in the scenes during each training step.
For the shape branch, we define a maximum batch size \( B^*_s \) and select scenes where the total number of objects \( B_s \) closely approaches but does not exceed this limit. This allows efficient use of batch capacity, though the batch size \( B_s \) fluctuates slightly due to varying object counts. Both \( B_l \) and \( B^*_s \) are set to 128 during training.
Additionally, when training the Visual Enhancement Module and the Relation Predictor module separately, the batch size is also set to 128.

\subsubsection{Training Procedure.}
The training process is divided into two distinct stages. In the first stage, we train the Visual Enhancement Module and the Relation Predictor separately. In the second stage, we train the dual-branch diffusion model.

The Visual Enhancement Module is trained on over 4,000 3D objects extracted from the SG-FRONT training set, where each object’s image information is sourced from the 3D-FRONT dataset. For each object, we use CLIP ViT-B/32 to extract textual and visual features, forming corresponding textual-visual pairs. 
The textual features are then quantized by selecting the 4 closest entries from a codebook \( \mathcal{C} \in \mathbb{R}^{64 \times 512} \), where the codebook consists of 64 entries, each with a dimension of 512.
The Visual Enhancement Module is trained for 1,000 epochs with a batch size of 128, which runs for 500 steps per epoch, employing the loss function \( \mathcal{L}_r \). The AdamW optimizer is used with a learning rate of \( 1 \times 10^{-4} \) and a weight decay of 0.02. Additionally, an exponential moving average (EMA) with a decay factor of 0.9999 is applied to stabilize the training process. This training strategy ensures consistent and robust learning of the module across the dataset.

The Relation Predictor uses training data generated by first masking 50\% of the text, image, and relationship information in the Full-Modality Graph, and then encoding the masked data into triplet representations.
The Relation Predictor model consists of a 10-layer GCN with a hidden dimension of 256, followed by two fully connected MLP layers with dimensions of 256 and 128, respectively. The model is trained using the loss function \( \mathcal{L}_e \), focusing on predicting the masked relationships.
Training proceeds for 1,000 epochs with a batch size of 128, using CrossEntropyLoss. AdamW is employed as the optimizer, with an initial learning rate of \( 5 \times 10^{-3} \).

The dual-branch diffusion model's training data is generated by applying a random masking ratio to the text and image of the Full-Modality Graph, which is subsequently encoded into LMMG. 
The model is trained using the loss function \( \mathcal{L}_o \) for 2050 epochs, with a batch size of \( B^*_s = 128 \) for the shape branch and \( B_l = 128 \) for the layout branch.
We utilize the AdamW optimizer, setting the initial learning rate to \( 1 \times 10^{-4} \).

\begin{figure}[t]
\centering
\includegraphics[width=0.45\textwidth]{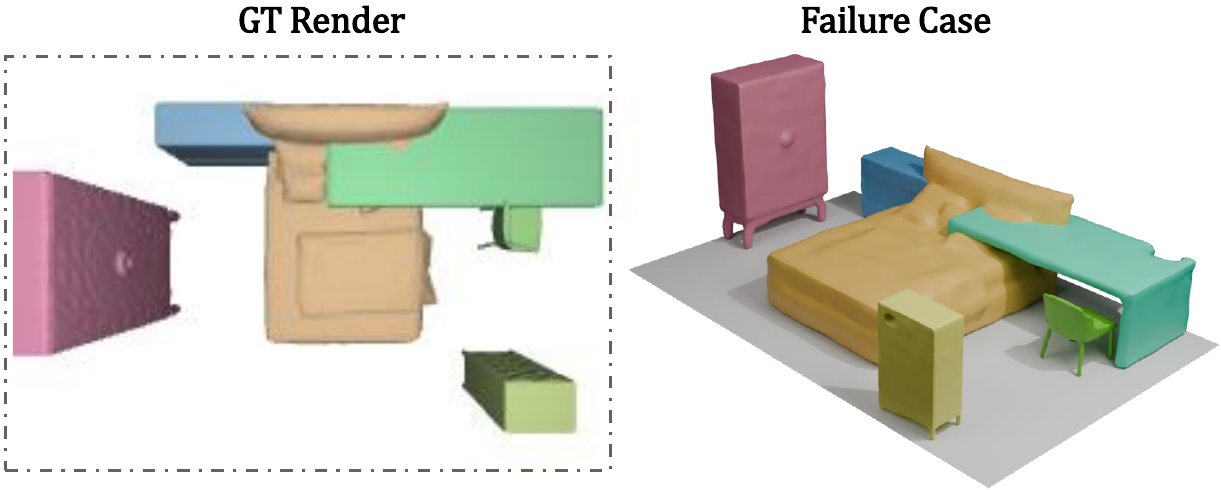} 
\caption{\textbf{Failure case.} The dashed box on the left is a top-down view rendered using the ground truth, while the result on the right is generated scene by MMGDreamer.}
\label{failure_case}
\end{figure}

\section{Limitations and Future Work}
\label{sec:Limitations and Future Work}
Our method demonstrates strong potential in generating complex 3D indoor scenes, yet it occasionally encounters failure cases, as shown in Fig.~\ref{failure_case}. These errors primarily stem from the limitations of the 3D-FRONT dataset, where noisy data often leads to interpenetrating objects in the generated scenes. While we implement post-processing techniques to minimize this noise, a small amount of erroneous data, such as overlapping furniture instances, remains in the dataset. This issue is reflected during inference, with some generated scenes showing minor collisions between objects. Nevertheless, these errors are infrequent, and our method consistently outperforms others in maintaining coherence between shape and layout despite the dataset's limitations.

While our method successfully integrates visual information, we have intentionally focused on generating objects with accurate geometric shapes and coherent scene layouts, deliberately excluding texture and material details for simplicity and control. Incorporating textures and material properties would add a new layer of complexity to the method, as modeling complex 3D shapes with detailed textures is a challenging task. Nevertheless, we recognize that including texture and material information presents an exciting opportunity for future work. By enhancing the method to better leverage visual data, we plan to generate scenes with richer texture details and achieve a higher degree of control over both geometry and texture, which will significantly improve the realism and versatility of our generated scenes.

\begin{figure*}[!t]
    \centering
    \begin{center}
    \begin{tcolorbox} [top=2pt,bottom=2pt, width=\linewidth, boxrule=1pt]
    {\footnotesize {\fontfamily{zi4}\selectfont
    \textbf{Mixed-Modality Graph Generation Prompt:}
  Assume you are an interior designer, and I will provide you with a multimodal scene design request that may include textual descriptions or images of furniture. \\
  Please create a graph based on my input and list all nodes along with the relationships between them (in the format A -> <relationship> -> B). Here are the constraints: \\
  1. For furniture described in text, the node name should be the corresponding English word. \\
  2. For furniture presented in images, you will first need to identify the type of furniture depicted in each image (Only need to identify its type without describing its attributes).  \\
  3. Additionally, number the images sequentially as Image1, Image2, etc., according to the order they were provided, and use the format "number (English word)" as the node name. When providing design requirements, focus solely on outlining the nodes and their relationships, without including any introductory or concluding remarks.\\
  
  Please note that only these twelve relationships are allowed: left of, right of, front, behind, close by, above, standing on, bigger than, smaller than, taller than, shorter than, symmetrical to, same style as, same super category as, same material as. When the input relationship description is not one of these twelve expressions, you need to replace it with a synonym from this list. \\
  
    *ONE-SHOT EXAMPLE* \\
    
    Here is an example of the output. Please make sure to output in this format: \\
    
    Nodes: \\
    - Image1 (Nightstand) \\
    - Image2 (Bed)   \\
    - Wardrobe   \\
    - Pendant Lamp  \\
    - Nightstand   \\
    
    Relationships:   \\
    - Image1 (Nightstand) -> close by -> Image2 (Bed) \\
    - Image1 (Nightstand) -> right of -> Pendant Lamp  \\
    - Wardrobe -> close by -> Image1 (Nightstand)  \\
    - Image1 (Nightstand) -> smaller than -> Image2 (Bed)   \\
    - Pendant Lamp -> behind -> Nightstand   \\
    - Image2 (Bed) -> front-> Wardrobe     \\
    - Pendant Lamp -> smaller than -> Image1 (Nightstand)   \\
    
    Here is my design requirement:
}
    \par}
    \end{tcolorbox}

    \end{center}
    \caption{\textbf{Prompt template} for Mixed-Modality Graph Generation with GPT-4V.}
    \label{fig:prompt-gpt}
\end{figure*}

\end{document}